%% file: main.tex
\documentclass{article}

\usepackage[final]{neurips_2025}

\usepackage[utf8]{inputenc} 
\usepackage[T1]{fontenc}    
\usepackage{hyperref}       
\usepackage{url}            
\usepackage{booktabs}       
\usepackage{amsfonts}       
\usepackage{nicefrac}       
\usepackage{microtype}      
\usepackage{xcolor}         
\usepackage{enumitem}
\usepackage{graphicx} 
\usepackage{amsmath}
\usepackage{float}

\title{HunyuanCustom: A Multimodal-Driven Architecture for Customized Video Generation}

\author{
    Tencent Hunyuan
}

\begin{document}

\maketitle

\vspace{-0.8cm}
\begin{figure}[H]
\includegraphics[width=1.0\textwidth]{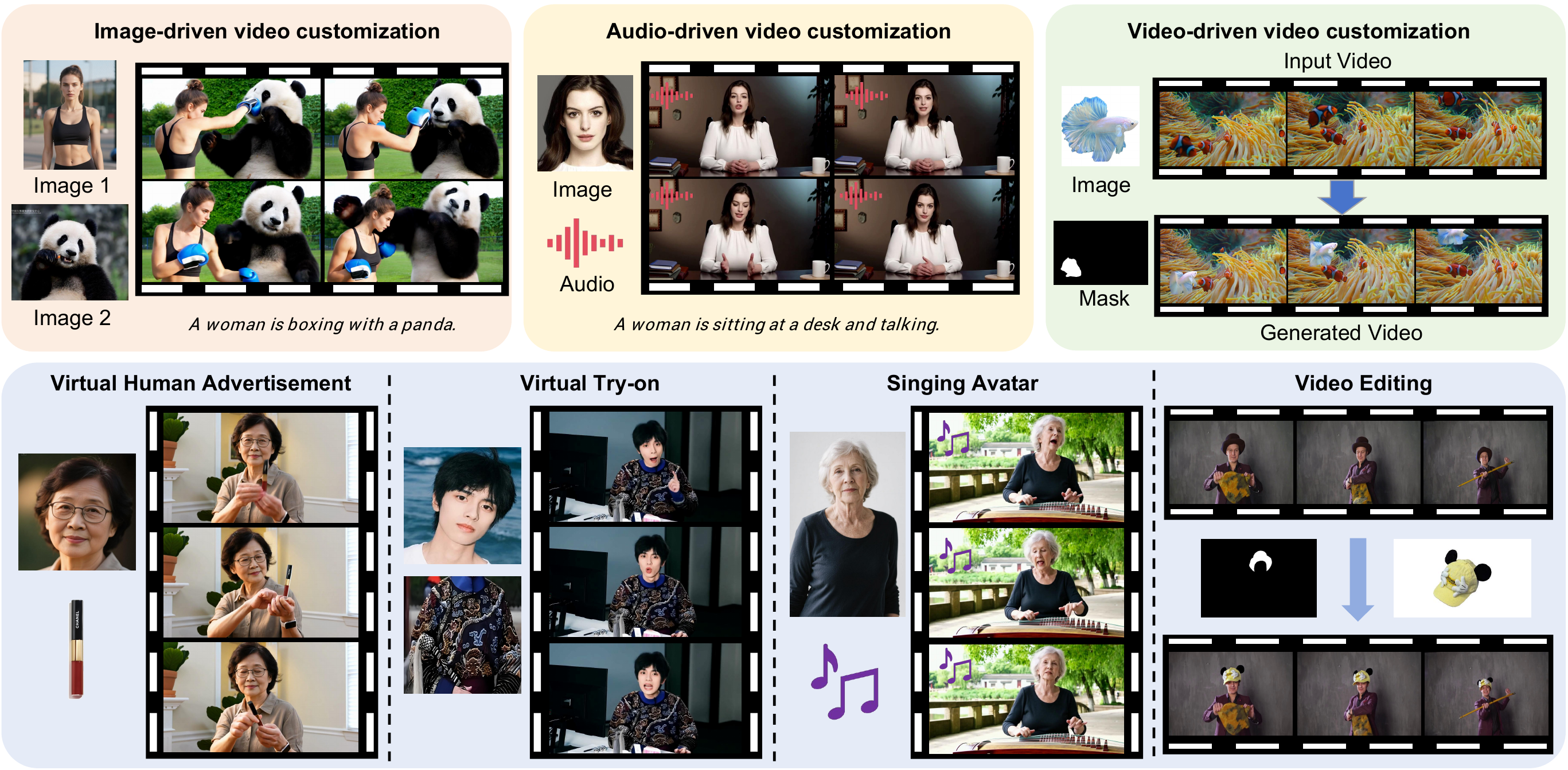}
\vspace{-0.3cm}
\caption{HunyuanCustom facilitates multi-modal driven video customization, allowing for the generation of videos based on text, images, audio, and video inputs. It supports a wide range of applications, such as virtual human advertisements, virtual try-ons, singing avatars, and video editing, significantly enhancing the controllability of subject-centric video generation.}
\vspace{-0.1cm}
\label{fig:teaser}
\end{figure}

\begin{abstract}

Customized video generation aims to produce videos featuring specific subjects under flexible user-defined conditions, yet existing methods often struggle with identity consistency and limited input modalities. In this paper, we propose HunyuanCustom, a multi-modal customized video generation framework that emphasizes subject consistency while supporting image, audio, video, and text conditions. Built upon HunyuanVideo, our model first addresses the image-text conditioned generation task by introducing a text-image fusion module based on LLaVA for enhanced multi-modal understanding, along with an image ID enhancement module that leverages temporal concatenation to reinforce identity features across frames. To enable audio- and video-conditioned generation, we further propose modality-specific condition injection mechanisms: an AudioNet module that achieves hierarchical alignment via spatial cross-attention, and a video-driven injection module that integrates latent-compressed conditional video through a patchify-based feature-alignment network. Extensive experiments on single- and multi-subject scenarios demonstrate that HunyuanCustom significantly outperforms state-of-the-art open- and closed-source methods in terms of ID consistency, realism, and text-video alignment. Moreover, we validate its robustness across downstream tasks, including audio and video-driven customized video generation. Our results highlight the effectiveness of multi-modal conditioning and identity-preserving strategies in advancing controllable video generation. All the code and models are available at \url{https://hunyuancustom.github.io}.
\end{abstract}


\input{secs/1_introduction}

\input{secs/2_related_work}

\input{secs/3_method}

\input{secs/4_experiment}

\input{secs/5_conclusion}

\bibliography{main}
\bibliographystyle{abbrvnat}
\end{document}

%% file: secs/1_introduction.tex
\section{Introduction}
\label{sec:introduction}
The field of video generation has undergone rapid advancement in recent years, driven by the proliferation of both open-source and commercial video-generation models. These advancements have significant real-world implications, ranging from content creation in the entertainment industry to applications in education, advertising, and more~\cite{xu2025hunyuanportrait,hu2024motionmaster,zhou2024avatargpt,pan2024expressive,huang2024make}. However, a critical limitation persists: the lack of precise controllability in current models. Generating videos that adhere to users’ specific requirements is still challenging, which restricts their potential applications in real-world scenarios where fine-grained customization is essential.

Controllable video generation often focuses on synthesizing videos featuring specific subjects, which is also known as \textbf{customized video generation}. Some existing methods such as, ConsisID~\citep{ConsisID} and MovieGen~\citep{moviegen} specialize in generating videos for a single human ID, they are unable to handle arbitrary objects. Other approaches, including ConceptMaster~\citep{huang2025conceptmaster}, Video Alchemist~\citep{video-alchemis}, Phantom~\citep{liu2025phantom}, and SkyReels-A2~\citep{fei2025skyreels} extend this capability to multi-subject generation. Nevertheless, these methods struggle with maintaining subject consistency and video quality, and their reliance on single-modality (image-driven) inputs restricts their broader applicability. Recently, VACE~\citep{jiang2025vace}, based on the \textit{Wan} video generation model~\citep{wang2025wan}, introduces a multi-modal-conditioned video generation framework. However, its excessive training tasks compromise ID consistency. Therefore, in this work we prioritize subject-consistent generation and develop a multi-modal customized video generation model that robustly preserves subject consistency. Our model supports diverse inputs, including image identities, audio conditions, video backgrounds, and text prompts, enabling multi-modal subject-consistent video generation.

In this work, we propose HunyuanCustom, a video generation model built upon HunyuanVideo which specializes in \textbf{subject-consistent generation conditioned on images, videos, audio, and text}. Specifically, our model first \textbf{generates videos that align with a given image indentity under text-driven condition}. We propose a text-image fusion module based on LLaVA, enabling interactive integration of text and images to enhance the model’s understanding of both modalities. Additionally, we propose an image ID enhancement module that leverages temporal concatenation of image information across video frames. By exploiting the video model’s inherent efficiency in time-series information transmission, this module effectively strengthens video ID-consistency.

Based on the subject-consistent customized video generation framework, HunyuanCustom extends its capabilities to audio and video modalities, enabling \textbf{audio-driven customized video generation and video-driven customized video generation}. To decouple the audio, video, and image modalities, HunyuanCustom employs distinct condition injection mechanisms for audio and video, ensuring independence from the image-level identity injection module. For audio-driven video customization, we propose AudioNet, which extracts multi-level deep audio features and injects them into corresponding video features via spatial cross-attention, achieving hierarchical audio-video alignment. For video-driven video customization, HunyuanCustom proposes an alignment and fusion module between conditional video and latent representations. By compressing the given video into the latent space through VAE, we project the video into the same space as the noisy latents. To compensate for the feature differences between clear video features and noisy latents, we design a video patchify module for video-latent feature alignment. Then, a new identity-disentangled video conditioning module is introduced to ensure seamless integration, enabling efficient video feature injection into the latent space.


HunyuanCustom has been rigorously evaluated on single-subject consistency and multi-subject consistency generation. We compared it with existing open-source methods and closed-source commercial software, conducting comprehensive comparisons across key metrics such as ID consistency, generation quality, and video-text alignment. The experimental results show that HunyuanCustom outperforms all existing methods in customized video generation. In addition, we validate its robustness through extensive experiments on audio and video-driven video customization, highlighting the superior performance of our method. Thanks to its strong identity preservation and multi-modal control capabilities, HunyuanCustom shows great potential for real-world applications such as virtual human advertising, virtual try-on, and fine-grained video editing. These results demonstrate the effectiveness of our HunyuanCustom, providing a solid foundation for future research in controllable subject-consistent video generation.

%% file: secs/2_related_work.tex
\section{Related Work}
\label{sec:related work}

\subsection{Video Generation Model}
Recent advancements in video generation have been significantly driven by diffusion models, which have successfully evolved from static image synthesis~\citep{rombach2022high,li2024hunyuan,flux2024} to dynamic spatio-temporal modeling~\citep{hong2022cogvideo,zhang2023i2vgen}.
The field has witnessed substantial progress with large-scale frameworks~\citep{liu2024sora,yang2024cogvideox,hunyuanvideo,wang2025wan,zhou2024allegro}, which demonstrate unprecedented high-quality content creation and a diverse array of generated results through extensive training on video-text pairs.
However, existing methods primarily concentrate on either text-guided video generation~\citep{lin2025mvportrait} or video generation based on a single reference image~\citep{gao2023high,xu2025hunyuanportrait}. These approaches often struggle to provide fine-grained control over the generated content and precise concept-driven editing. This limitation continues to exist despite advancements in multi-condition control. 
While pioneering work such as VACE~\citep{jiang2025vace} enables multi-condition capabilities through multi-modal modeling, it fails to maintain identity consistency due to the excessive number of training tasks. In this study, we meticulously design a multi-condition-driven model that incorporates various modalities, including images, videos, audios, and texts, while also emphasizing subject-consistency generation.

\subsection{Video Customization}

\textbf{Instance-specific video customization.} Instance-specific video customization methods~\citep{wu2025customcrafter,wang2024customvideo,idanimator} that take several images with the same identity to fine-tune the pretrained video generation model, where each identity is trained separately. To capture the identity information of target images, Textual Inversion~\citep{textual_inversion} and DreamBooth~\citep{ruiz2023dreambooth} embed image identity information into the text space, enabling effective interaction with text and facilitating image generation of the target identity under the corresponding text. To extend these methods to video generation, Still-Moving~\citep{stillmoving} proposes first fine-tuning a video generation model based on PEFT methods to generate videos of static frames, then repeating the image as a static video and using DreamBooth to learn the target identity. Similarly, CustomCrafter~\citep{wu2025customcrafter} repeats the image for n frames and then embeds it into the text space using textual inversion, further fine-tuning the video generation model to better learn the target identity. CustomVideo~\citep{wang2024customvideo} extends customization from a single subject to multiple subjects by segmenting multiple subject images and stitching them into one, binding the subject identity to the corresponding text through alignment with the cross-attention map and subject mask. DisenStudio~\citep{chen2024disenstudio} associates different subjects with attention regions, ensuring each subject appears only in specific locations within the video, thereby assigning different actions to different subjects. However, these methods rely on instance-specific optimization, which poses a challenge in real-time or large-scale video customization.

\textbf{End-to-end video customization.} End-to-end video customization methods inject the identity information of the target images into the video generative models by training an additional conditioning network, allowing generalization to arbitrary identity image inputs during the inference stage, significantly advancing video customization. Some previous works focus on maintaining facial identity~\citep{idanimator,ConsisID,moviegen}. For instance, ID-Animator~\citep{idanimator} introduces a face adapter and incorporates a facial identity loss to maintain facial ID consistency. ConsisID~\citep{ConsisID} extracts low-frequency and high-frequency information from facial images through global and local facial extractors to achieve comprehensive ID capture. MovieGen~\citep{moviegen} injects facial ID information into the text space and uses facial images from different videos to guide video generation, alleviating the issue of facial copying. To achieve customization of arbitrary objects, VideoBooth~\citep{jiang2024videobooth} injects identity information of any image through coarse-grained CLIP features and fine-grained image features. Recently, some works including ConceptMaster~\citep{huang2025conceptmaster}, Video Alchemist~\citep{video-alchemis}, Phantom~\citep{liu2025phantom}, SkyReels-A2~\citep{fei2025skyreels}, and VACE~\citep{jiang2025vace} have extended customization from a single subject to multiple subjects by binding words in text prompts to corresponding subject images, enabling the generation of videos with multiple subjects. However, due to the mutual influence between multiple IDs and the complexity of interactions among multiple subjects, there remains significant room for improvement in maintaining and interacting with multiple subject IDs.

%% file: secs/3_method.tex
\section{Method}
\label{sec:method}
\subsection{Overview}
\label{sec:overview}

HunyuanCustom is a multi-modal customized generation model centered on subject consistency, built upon the Hunyuan Video generation framework~\citep{li2024hunyuan}. It enables the generation of subject-consistent videos conditioned on text, images, audio, and video inputs, as shown in Fig.~\ref{fig:main framework}. Specifically, HunyuanCustom introduces an image-text fusion module based on LLaVA to facilitate interaction between images and text, allowing identity information from images to be effectively integrated into textual descriptions. Additionally, an identity enhancement module is proposed, which concatenates image information along the temporal axis and leverages the video model's efficient temporal modeling ability to enhance subject identity throughout the video. To support conditional injection of audio and video, HunyuanCustom designs distinct injection mechanisms for each modality, which are effectively disentangled with the image-level identity condition module. HunyuanCustom ultimately achieves decoupled control over image, audio, and video conditions, demonstrating great potential in subject-centric multi-modal video customization.

\subsection{Multi-modal task}
\label{sec:multi-modal task}
HunyuanCustom supports the conditions from text, image, audio, and video. All the tasks are built upon the ability to generate ID-consistent videos. As shown in Fig.~\ref{fig:main framework}. The tasks can be classified into the following 4 categories:

\begin{figure}[t]
\centering
\includegraphics[width=1.0\textwidth]{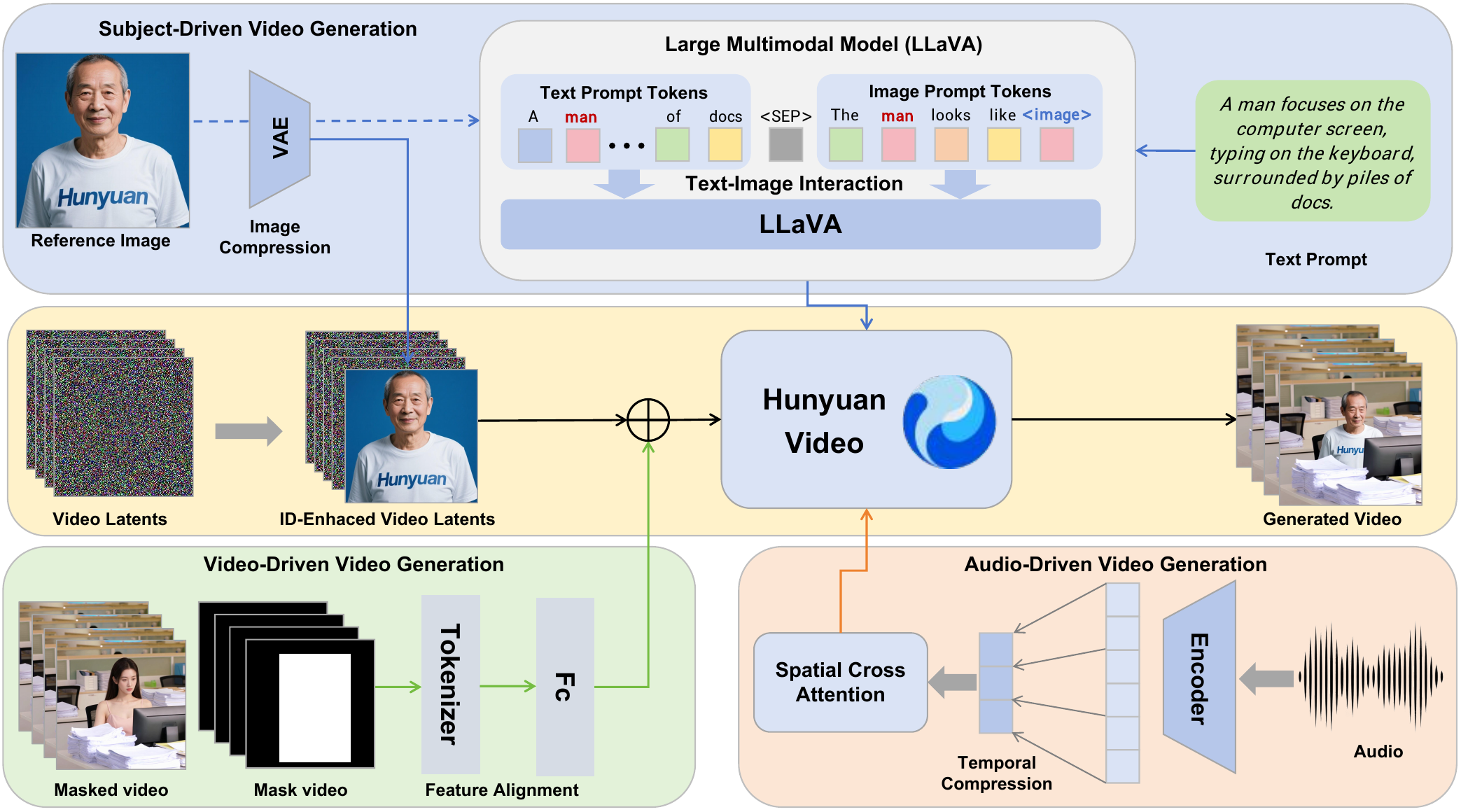}
\caption{The main framework of HunyuanCustom, where we can generate identity-consistent videos conditioned on text, image, audio, and video.}
\label{fig:main framework}
\end{figure}

\begin{itemize}[leftmargin=*,noitemsep,nolistsep]
    \item \textbf{Text-driven video generation.} The text-to-video generation ability comes from the base model HunyuanVideo, which supports generating videos aligned with the given text prompts;
    \item \textbf{Image-driven video customization.} At the core of HunyuanCustom is the ability to take an input image, extract identity information, and generate a corresponding video of that identity guided by a textual description—enabling \textit{customized video generation}. HunyuanCustom supports both human and non-human identities, and further allows multiple identities as input, enabling interactive generation involving multiple subjects.
    \item \textbf{Audio-driven video customization.} Building on subject customization, HunyuanCustom incorporates audio as an additional modality. Given a human identity, a text prompt, and corresponding audio, the system enables the subject to perform actions synchronized with the audio (e.g., speaking, playing, singing) within the context described by the text. This extends conventional audio-driven human animation by allowing the specified identity to perform freely in arbitrary scenes and actions, significantly enhancing controllability.
    \item \textbf{video-driven video customization.} HunyuanCustom also supports video-to-video generation by enabling object replacement or insertion based on identity customization. Given a source video and an image specifying the target identity, the system can replace an object in the video with the specified identity. Additionally, it allows inserting the identity into a background video based on textual guidance, enabling flexible object addition.
\end{itemize}

\subsection{Multi-Modal Data Construction}

\begin{figure}[t]
    \centering
    \includegraphics[width=1.0\textwidth]{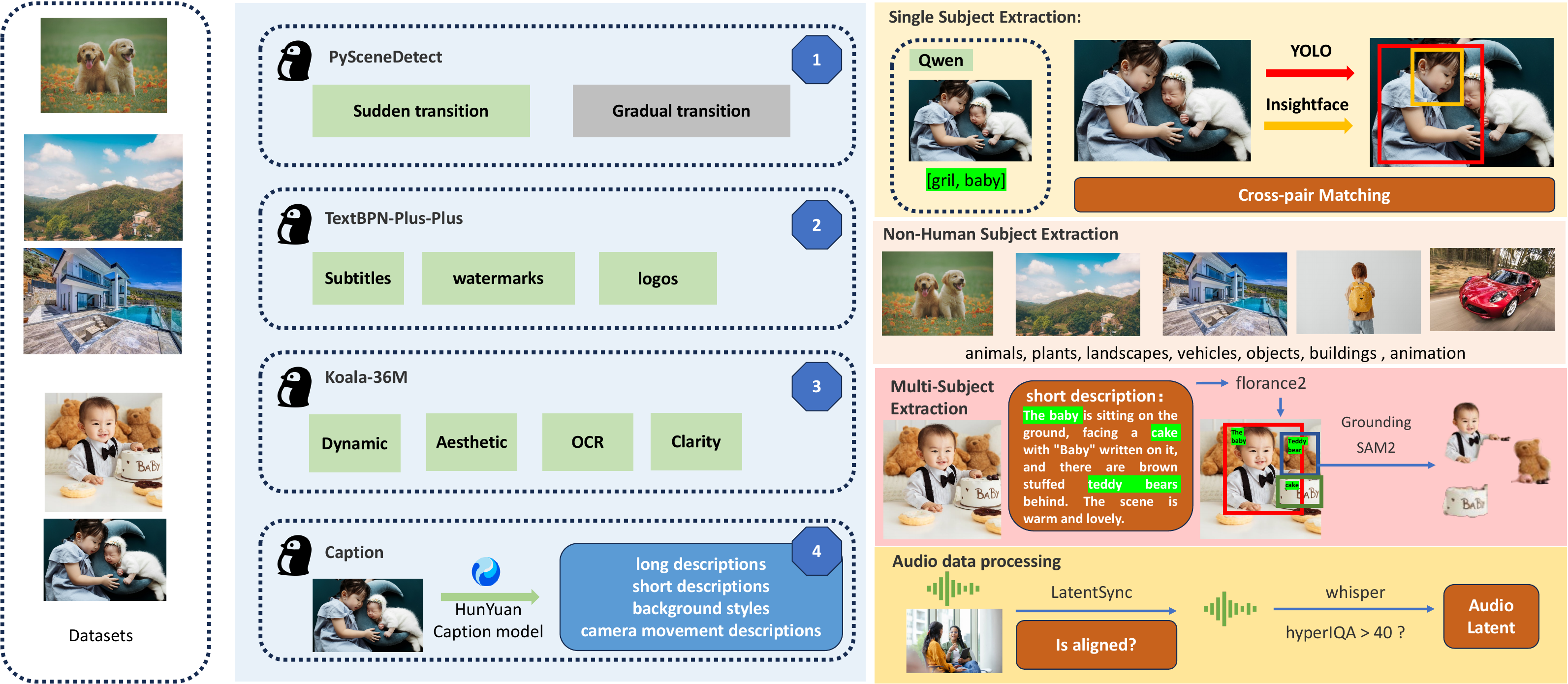}
    \caption{Data Construction Pipeline.}
    \label{fig:Data Construction Pipeline}
\end{figure}

Our data undergoes a rigorous processing pipeline to ensure high-quality inputs that enhance model performance. Experimental results demonstrate that high-quality data plays a crucial role in tasks such as subject consistency, video editing, and audio-driven video generation. While different tasks may follow their own specific data processing steps, the initial processing stages are common across tasks, with the key differences lying in the subsequent steps. In light of this, this section delves into the detailed methodologies of video data preparation, focusing on the shared preprocessing techniques as well as the task-specific post-processing approaches designed for distinct tasks.

Our data is sourced from diverse channels, and to ensure strict compliance with the principles outlined in the General Data Protection Regulation (GDPR) \citep{regulation2018general} framework, we employ data synthesis and privacy-preserving computation techniques to regulate the data collection process. The raw data spans a wide range of domains, primarily encompassing eight major categories: humans, animals, plants, landscapes, vehicles, objects, architecture, and anime. In addition to our self-collected data, we have rigorously curated and processed open-source datasets (e.g., OpenHumanvid~\citep{li2024openhumanvid}), which significantly expand the diversity of our data distribution and enhance model performance. Experimental results confirm that the incorporation of high-standard data is crucial for achieving substantial improvements in model performance.

\textbf{Data Filtering and Preprocessing. }
Given the broad distribution of our dataset, which also includes open-source data, there are significant variations in duration, resolution, and quality among the videos. To address these issues, we implemented a series of preprocessing techniques. Firstly, to prevent transitions within training data, we utilized PySceneDetect~\citep{castellano2020pyscenedetect} to segment the original videos into single-shot clips. For handling text regions in videos, we employed textbpn-plus-plus~\citep{zhang2023arbitrary} to filter out clips with excessive text and cropped videos containing subtitles, watermarks, and logos. Due to the uneven distribution of video sizes and durations, we performed cropping and alignment, standardizing the short side to either 512 or 720 pixels and limiting video length to 5 seconds (129 frames). Finally, considering that PySceneDetect cannot detect gradual transitions and textbpn-plus-plus~\citep{zhang2023arbitrary} has limited capability in detecting minor text, and to ensure aesthetic quality, motion magnitude, and scene brightness, we used the koala-36M~\citep{wang2024koala} model for further refinement. However, due to differences between the training data of koala-36M~\citep{wang2024koala} and our dataset, and its lack of fine-grained assessment on aesthetic quality and motion magnitude, we established our own evaluation criteria, determining a koala threshold of 0.06 specific to our dataset for meticulous filtering. Experimental results confirm the importance of our data selection and processing methods in enhancing model performance.


\textbf{Subject Extraction. }
Single Subject Extraction: To extract the main subject from videos, we first use the Qwen7B~\citep{bai2023qwen} model to label all subjects in each frame and extract their IDs. Subsequently, we employ a clustering algorithm (e.g., Union-Find) to compute the frequency of each ID's appearance across frames and select the ID with the highest occurrence as the target subject. Multiple IDs can be chosen if necessary; however, if all IDs appear fewer than a predefined threshold (e.g., 50 frames), the video is discarded. Next, we use YOLO11X~\citep{khanam2024yolov11} for human body segmentation to obtain bounding boxes and InsightFace~\citep{ren2023pbidr} to detect face positions and generate face bounding boxes. If the proportion of the face bounding box within the human body bounding box is less than 0.5, the detection result from YOLO11X is considered erroneous, and the corresponding bounding box is discarded.

Non-Human Subject Extraction: For non-human subjects, we utilize QwenVL~\citep{bai2025qwen2} to extract subject keywords from the video and employ GroundingSAM2~\citep{ravi2024sam2segmentimages, liu2023grounding, ren2024grounding, ren2024grounded, jiang2024trex2} to generate masks and bounding boxes based on these keywords. If the size of a bounding box is less than 0.3 times the dimensions of the source video, it is discarded. To ensure balanced category distribution in the training data, we use QwenVL to classify the main subject into one of eight predefined categories: animals, plants, landscapes, vehicles, objects, architecture, and anime. We then apply balanced sampling across these categories to achieve an equitable distribution.

Multi-Subject Extraction: For multi-subject scenarios, we use QwenVL to filter videos from single-person datasets that involve interactions between humans and objects. Since we need to align the subject keywords in video captions with those in images, directly using QwenVL to re-extract subject keywords may lead to misalignment with the keywords in the video prompt. Therefore, we employ Florence2~\citep{xiao2024florence} to extract bounding boxes for all subjects mentioned in the video captions. Subsequently, GroundingSAM2 is used to perform subject extraction on these bounding box regions. We then apply clustering to remove frames that do not contain all subjects. To address issues related to hard-copying, we use the first 5 seconds of the video for model training and the subsequent 15 seconds for subject segmentation.

\textbf{Video Resolution Standardization. }
We first compute a union bounding box based on all the bounding boxes of the main subjects and ensure that the cropped region contains at least 70\% of the area of the union bounding box. To enable the model to support multi-resolution outputs, we define several aspect ratios, including 1:1, 3:4, and 9:16.

\textbf{Video Annotation. }
We employ a structured video annotation model developed by the HunYuan team to label the videos. This model provides detailed descriptive information, including long descriptions, short descriptions, background styles, and camera movement descriptions of the videos. During the training process, these structured annotations are utilized to enhance the video captions, thereby improving the robustness and performance of the model.

\textbf{Mask Data Augmentation. }
During video editing, directly using the extracted subject masks for training can lead to overfitting when replacing objects of different types or shapes. For instance, replacing a doll without ears with one that has ears might result in the generated video still showing the doll without ears, which is not the desired outcome. Therefore, during the training process, we apply techniques such as mask dilation or converting masks to bounding boxes to soften the mask boundaries. These methods help achieve more realistic and expected editing results in the final video. By employing these augmentation strategies, we aim to mitigate overfitting issues and ensure that the edited videos meet our expectations more closely. This approach enhances the flexibility and applicability of the model across various object types and shapes.

\textbf{Audio data processing.}
First, we utilize LatentSync~\citep{li2024latentsync} to evaluate the synchronization between audio and video in the clips. Specifically, we discard videos with a synchronization confidence score below 3 and adjust the audio-video offset to zero. At the same time, we compute the hyperIQA quality score and remove any videos scoring below 40 to ensure high-quality data. Finally, we employ Whisper~\citep{radford2023robust} to extract audio features, which will be used as input for subsequent model training.

\subsection{Image-driven Video Customization}


At the core of HunyuanCustom is the task of generating videos conditioned on an input image $I$ representing a specific identity and a textual description $T$. A key challenge lies in enabling the model to effectively comprehend the identity information embedded in the image and integrate it with the textual context for interactive understanding. To this end, HunyuanCustom incorporates a LLaVA-based text-image interaction module, which facilitates joint modeling of visual and textual inputs, thereby enhancing the model’s understanding of both the identity and the accompanying description. Additionally, an identity enhancement module is introduced, which propagates image-derived features across the video sequence. This leverages the video model’s temporal modeling capabilities to reinforce identity consistency throughout the generated video.

\textbf{LLaVA-based text-image interaction.} In the context of video customization, effectively integrating image-text information has been a key challenge for previous customization methods~\citep{fei2025skyreels,jiang2025vace,liu2025phantom}. These methods either lack a design for interactive understanding between image and text or rely on additional newly trained branch networks to achieve this interaction. HunyuanCustom leverages the text comprehension capabilities trained in the LLaVA~\citep{llava} text space by Hunyuan Video~\citep{hunyuanvideo} and utilizes LLaVA's inherent multimodal interaction understanding ability. By extending the original text input of HunyuanVideo to include both image and text inputs, HunyuanCustom achieves effective image-text interaction understanding based on LLaVA's outstanding multimodal comprehension abilities.

Specifically, given a text input $T$ and an image input $I$ with a corresponding description word $T_{I}$ in the text, we design a template to facilitate interaction between the text and image. We explore two types of templates: (1) the image-embedded template, where the description word $T_{I}$ in the text is replaced with the image token $\texttt{<image>}$ (e.g., for the text prompt "A man is playing guitar," if we input the identity image for "man," the resulting template is "A $\texttt{<image>}$ is playing guitar"); and (2) the image-appended template, where the image token is placed after the text prompt by adding an identity prompt, "The $T_{I}$ looks like $\texttt{<image>}$" (e.g., for the text prompt "A man is playing guitar," the resulting template is "A man is playing guitar. The man looks like $\texttt{<image>}$"). After processing, the image token $\texttt{<image>}$ is replaced by $24 \times 24$ image hidden features extracted by LLaVA. Since the image feature tokens are significantly longer than the text feature tokens, to prevent the image features from overly influencing text comprehension, we insert a special token $\texttt{<SEP>}$ between the text prompt and the image prompt. This helps the LLaVA model retain the information from the text prompt while establishing a connection between the text prompt and the image identity.

\textbf{Identity Enhancement.} The LLaVA model, as a multi-modal understanding framework, is designed to capture the correlation between text and image, primarily extracting high-level semantic information such as category, color, and shape, while often overlooking finer details like text and texture. However, in video customization, identity is significantly determined by these image details, making the LLaVA branch alone insufficient for identity preservation. To address this, we propose an identity enhancement module. By concatenating video latents with the target image over the time axis, and leveraging the video model's efficient information transmission capability in the temporal dimension, we can effectively enhance video identity consistency.

Specifically, we first resize the image to match the video frame size. We then employ the pretrained causal 3DVAE from Hunyuanvideo to map the image $I$ from image space to latent space. With the image latent $z_I \in \mathbb{R}^{wh \times c}$, where $wh$ represents the width and height of the latent and $c$ is the feature dimension, we concatenate the noisy video latent $z_t \in \mathbb{R}^{fwh \times c}$ (where $f$ is the number of video frames) and the image latent $z_I$ along the first sequence dimension to obtain a new latent $z = \{z_I, z_t\} \in \mathbb{R}^{(f+1)wh \times c}$. Given the pretrained Hunyuanvideo's strong prior in modeling temporal information, identity can be efficiently propagated along the time axis. Consequently, we assign the concatenated image latent with a 3D-RoPE~\citep{rope} along the time series. In the original Hunyuan video, the video latent is assigned a 3D-RoPE along the time, width, and height axes; for a pixel located at $(f, i, j)$ (where $f$ is the frame index, $i$ is the width, and $j$ is the height), it receives a RoPE with $RoPE(f, i, j)$. For the image latent, to enable effective identity broadcasting along the time series, we position the image latent at the $-1$-th frame, preceding the first frame with time index $0$. Furthermore, inspired by Omnicontrol~\citep{ominicontrol} in controllable image generation, to prevent the model from simply copying and pasting the target image into the generated frames, we introduce a spatial shift for the image latents, where:

\begin{equation}
        RoPE_{z_I}(f,i,j)=RoPE(-1,i+w,j+h).
\end{equation}

\textbf{Multi-subject Customization.} For multi-subject customization, we utilize the trained single-subject customization model as a foundation and subsequently fine-tune it to accommodate the multi-subject customization task. Specifically, we have several condition images $\{I_1, I_2, \ldots, I_m\}$, each with corresponding text descriptions $\{T_{I,1}, T_{I,2}, \ldots, T_{I,m}\}$. For each image, we template them as "the $T_{I,k}$ looks like \texttt{<image>}" and model the text-image correlation using the LLaVA model. Additionally, to enhance image identity, we encode all images into latent space to obtain image latents $\{z_{I,1}, z_{I,2}, \ldots, z_{I,m}\}$ using 3D-VAE, and then concatenate them with the video latent. To differentiate between various identity images, we assign the $k$-th image a time index of $-k$, which is associated with a 3D-RoPE:
\begin{equation}
        RoPE_{z_{I,k}}(f,i,j)=RoPE(-k,i+w,j+h).
\end{equation}

\textbf{Training Process.} In the training process, we adopt the Flow Matching~\citep{flowmatching} framework to train the video generation models. For training, we first acquire the video latent representation $z_1$ and the corresponding identity image $I$. Then, we sample $t\in[0, 1]$ from a logit-normal distribution~\citep{esser2024scaling} and initialize the noise $z_0\sim N(0, I)$ according to the Gaussian distribution. After that, we construct the training sample $z_t$ through linear interpolation. The model aims to predict the velocity $u_t = \frac{dz_t}{dt}$ conditioned on the target image $I$, which is used to guide the sample $z_t$ towards $z_1$. The model parameters are optimized by minimizing the mean-squared error between the predicted velocity $v_t$ and the real velocity $u_t$, and the loss function is defined as:

\begin{equation}
\mathcal{L}_{generation}=\mathbb{E}_{t, x_{0}, x_{1}}\left\|v_{t}-u_{t}\right\|^{2}.
\end{equation}

To endow our model with a more extensive representational capacity and enable it to capture and learn a broader range of complex patterns, we fully fine-tune the weights of both the pretrained video generation model and the LLaVA model, ultimately unlocking its full potential for delivering superior video customization results.


\section{Multimodal subject-centric video generation}
Previous video customization methods~\citep{ConsisID,huang2025conceptmaster,fei2025skyreels,liu2025phantom} primarily focus on maintaining subject identity, lacking further exploration into subject-driven video generation. We further investigate the use of multimodal audio-video information as a condition, centering on the subject identity, to achieve image-audio-video jointly driven subject-specific generation.

\subsection{Audio-driven video customization}

\textbf{Audio-driven video customization.} Audio is an indispensable component in video generation, with extensive research dedicated to using audio as a condition to drive video creation. Among these, \textit{\textbf{audio-driven human animation}} represents an important research topic. Existing models~\citep{jiang2024loopy,ji2024sonic} for audio-driven human animation typically use a human image and audio as input to animate the character in the image to speak the corresponding speech. However, this image-to-video paradigm results in generated videos where the character's posture, attire, and setting remain consistent with the input image, limiting the ability to generate videos of the target character in different postures, attire, and settings. This limitation restricts their application. Leveraging HunyuanCustom's effective capture and maintenance of character identity information, we further integrate audio input to enable the generation of videos where the character speaks the corresponding audio in a text-described scene, allowing for more flexible and controllable speech-driven digital human generation, which we call it \textit{\textbf{audio-driven video customization}}.


\textbf{Identity-disentangled AudioNet.}
To effectively decouple audio signals from identity information, we propose \textbf{Identity-disentangled AudioNet}. As described in Section~3, identity information is primarily injected via the text modality through LLaVA and through token concatenation along the latent temporal dimension. To ensure disentangled audio conditioning, AudioNet adopts an alternative conditioning mechanism that avoids entanglement with identity cues.
Given an audio-video sequence of $ f' $ frames, we first extract audio features for each audio frame, resulting in a tensor of size $ f' \times 4 \times c $, where 4 represents the number of tokens per audio frame. Since the video latent representation is temporally compressed by the VAE to $ f $ frames ($ f = \left\lfloor \frac{f'}{4} \right\rfloor + 1 $, where 1 corresponds to the uncompressed initial frame and 4 is the temporal compression ratio), the ID-enhanced video latent includes $ f + 1 $ frames due to the insertion of the identity image at the beginning.
To temporally align the audio features with the compressed video latent, we first pad the audio features before the initial frame to match $(f + 1) \times 4 $ frames. We then aggregate every 4 consecutive audio frames into a single frame, forming a new audio feature tensor $f_A $ that is temporally aligned with the video latent representation.

\begin{equation}
    \begin{aligned}
        f_A=Rearrange(f_{A,0}): [b,(f+1)\times4,4,c]\rightarrow [b,(f+1),16,c]
    \end{aligned}
\end{equation}

With the temporally aligned audio feature $f_A$, we employ a cross-attention module to inject audio information into the video latent $z_t$. To prevent inter-frame interference between audio and video at different frames, we adopt a \textbf{spatial cross-attention} mechanism that performs audio injection on a per-frame basis. Specifically, we decouple the temporal dimension from the spatial dimensions  in the video latent and apply cross-attention only along the spatial axes  (width $w$ and height $h$):

\begin{gather}
        z'_{t,A}=Rearrange(z_t):[b,(f+1)wh,c]\rightarrow [b,f+1,wh,c]\\\notag
        z''_{t,A}=z'_{t,A}+\lambda_A\times CrossAttn(f_A,z'_t)\\
        z_{t,A}=Rearrange(z''_{t,A}):[b,f+1,wh,c]\rightarrow [b,(f+1)wh,c]\notag
\end{gather}
where $\lambda_A$ is a weight to control the influence of the audio feature.

\subsection{Video-driven video customization}

In practical video creation, editing is a fundamental task that often involves modifying the appearance and motion of subjects within a video. This aligns naturally with HunyuanCustom’s subject generation capabilities, enabling subject-level editing such as replacement and insertion.
Videos contain rich spatiotemporal information, presenting challenges in both effective content extraction and efficient integration into the generative model. Existing methods, such as VACE~\citep{jiang2025vace}, inject video conditions via adapter modules, which double the computational cost and severely limit efficiency. Other approaches~\citep{bai2025recammaster} concatenate the video latents of the conditioning and generated clips along the temporal axis, leading to a doubled sequence length and quadratic growth in attention computation.
To overcome these limitations, HunyuanCustom adopts a more efficient video condition injection strategy that decouples video information from image and audio modalities. Specifically, it first compresses the conditioning video using the pretrained causal 3D-VAE, aligns the resulting features with the noisy video latents via feature alignment, and then directly adds the aligned features to the video latent representation. This enables efficient and effective incorporation of video conditions without incurring significant computational overhead.

\textbf{Video-Latent Feature Alignment.} The conditioning video serves as a clean, noise-free input, whereas the video latents are obtained from a noisy encoding process. To improve video condition injection, we first perform feature alignment between the conditioning video and the video latents. Specifically, the conditioning video is encoded using the pretrained causal 3D-VAE encoder, followed by compression and serialization via the pretrained video tokenizer in HunyuanVideo. A four-layer fully connected network then maps the conditioning video features into the latent space, achieving alignment with the video latents.

\textbf{Identity-Disentangled Video Conditioning.} We explore two strategies for injecting the conditioning video into the pretrained video generation model. The first concatenates the conditioning video features with the video latents along the token dimension, followed by dimensionality compression to project the result back into the original latent space. The second directly adds the conditioning video features to the video latents on a frame-by-frame basis along the temporal dimension, preserving the original feature dimensions. In both cases, the conditioned latents retain the same shape as the original video latents, introducing no additional computational overhead during inference.
Our experiments show that the concatenation-based method struggles to preserve content information and suffers from substantial information loss. In contrast, the addition-based method enables more effective content injection. Thanks to the prior feature alignment step, the conditioning video features and video latents are well-matched, facilitating efficient fusion and information transfer, and thereby supporting effective and lightweight video condition injection.

%% file: secs/4_experiment.tex
\section{Experiment}
\label{sec:experiment}

\begin{figure}[t]
    \centering
    \includegraphics[width=0.48\textwidth]{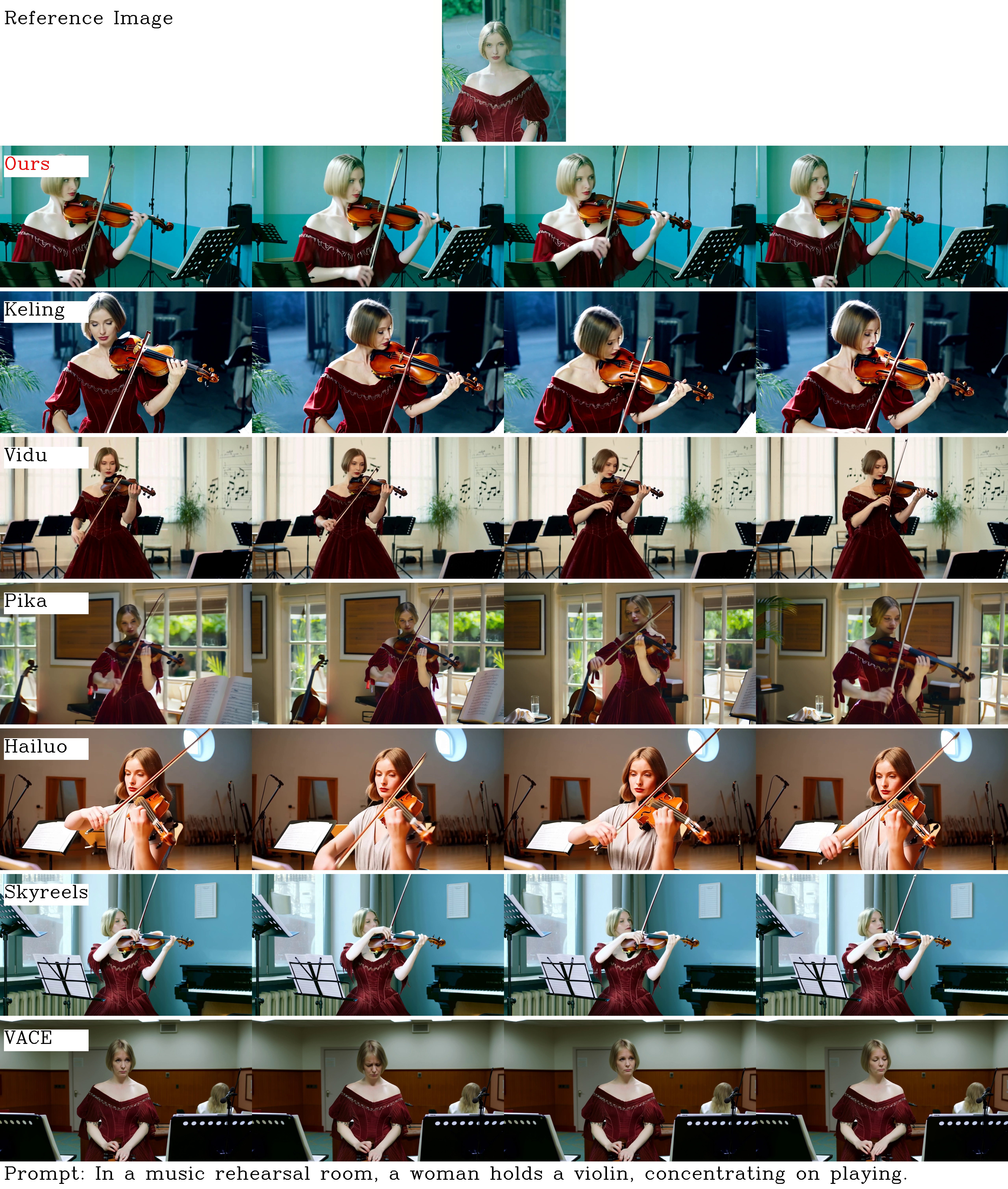}
    \hspace{0.002in}
    \includegraphics[width=0.48\textwidth]{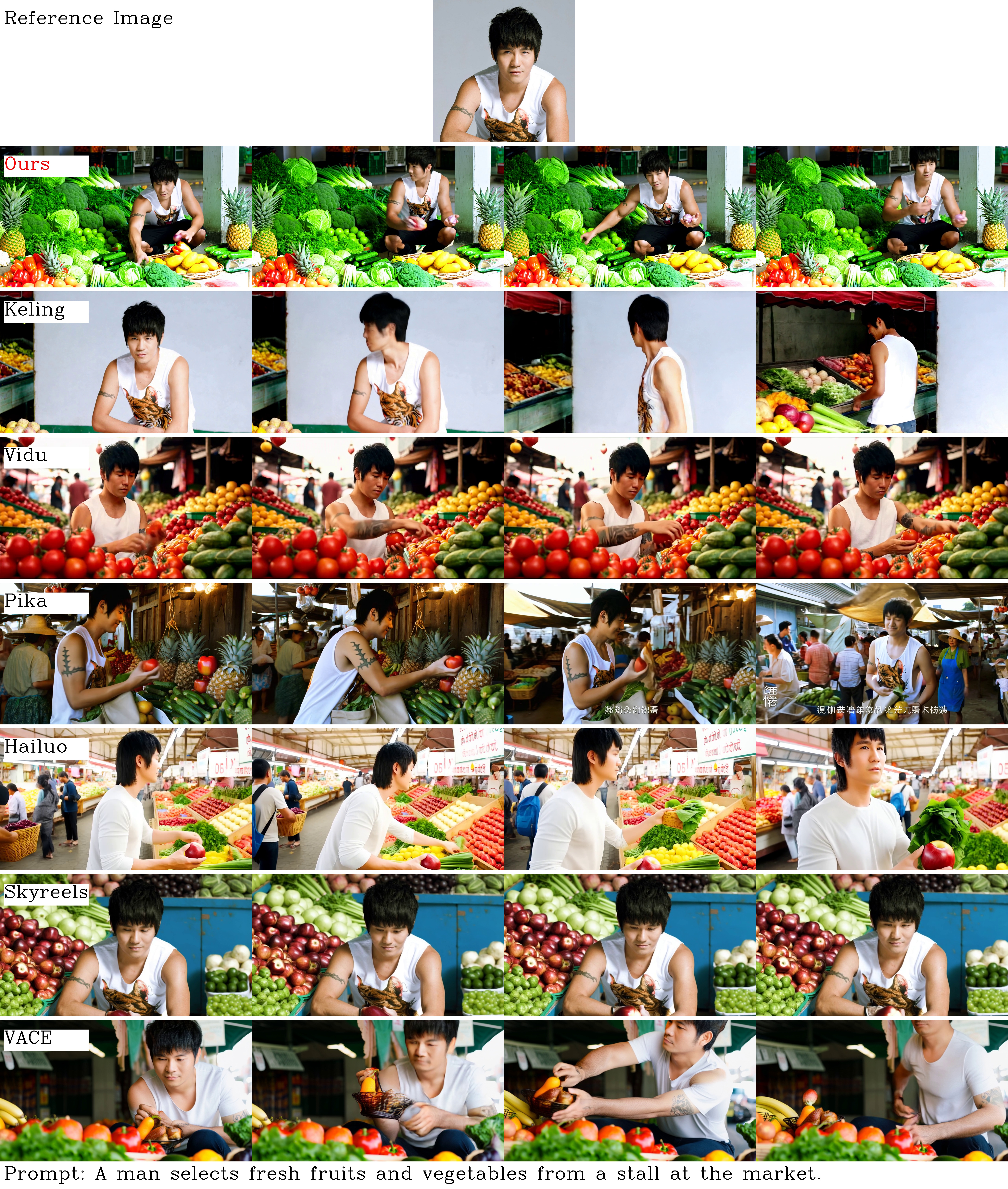}
    \caption{Comparison on human-centered video customization.}
    \label{fig:singleref_1}
\end{figure}

\subsection{Implementation details}

\textbf{Evaluation metrics.}To evaluate the performance of video customization, we employ the following metrics to evaluate the identity preservation, text-video alignment, and video generation quality:
\begin{itemize}
    \item \textbf{ID consistency.} We employ Arcface~\citep{deng2019arcface} to detect and extract the embedding of the reference face and each frames of generation video, and then compute the average cosine similarity between them.
    \item \textbf{Subject similarity.} First, we detect each frame and get the segment result of human using YOLOv11~\citep{khanam2024yolov11}, and then compute the similarity of the DINO-v2~\citep{oquab2023dinov2} feature between the reference and results.
    \item \textbf{Text-video alignment.} We employ CLIP-B~\citep{clip} to evaluate the alignment between the given text prompt and the corresponding generated videos.
        \item \textbf{Temporal consistency.} Following VBench~\citep{huang2024vbench}, we utilize the CLIP-B~\citep{clip} model to calculate the similarity between each frame and its adjacent frames, as well as the first frame, to assess the temporal consistency of the video.
    \item \textbf{Dynamic degree.} The dynamic degree is used to measure the movement of an object, which is calculated following VBench~\citep{huang2024vbench}.

\end{itemize}

\subsection{Comparison on single-subject video customization}

\textbf{Baselines.} We compare HunyuanCustom with the state-of-the-art video customization methods, including commercial products (Vidu 2.0~\citep{vidu}, Keling 1.6~\citep{Keling}, Pika~\citep{pika}, and Hailuo~\citep{Hailuo}) and open-sourced methods (Skyreels-A2~\citep{fei2025skyreels} and VACE~\citep{jiang2025vace}). For each model, we generate 100 videos with human identities and 100 videos with nonhuman identities, which can comprehensively demonstrate the general video customization ability of these methods. 

\begin{figure}[t]
    \centering
    \includegraphics[width=0.48\textwidth]{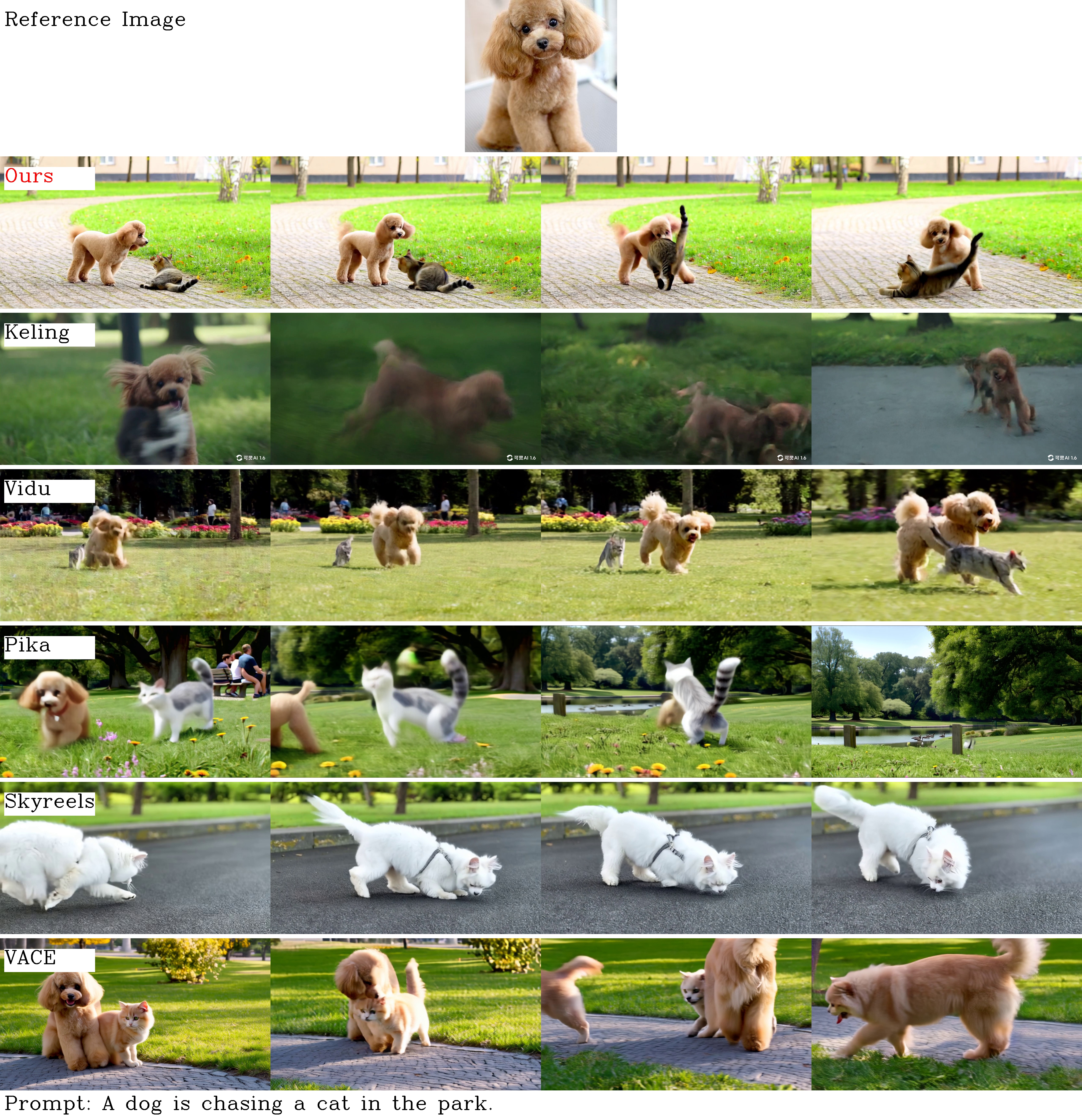}
    \hspace{0.002in}
    \includegraphics[width=0.48\textwidth]{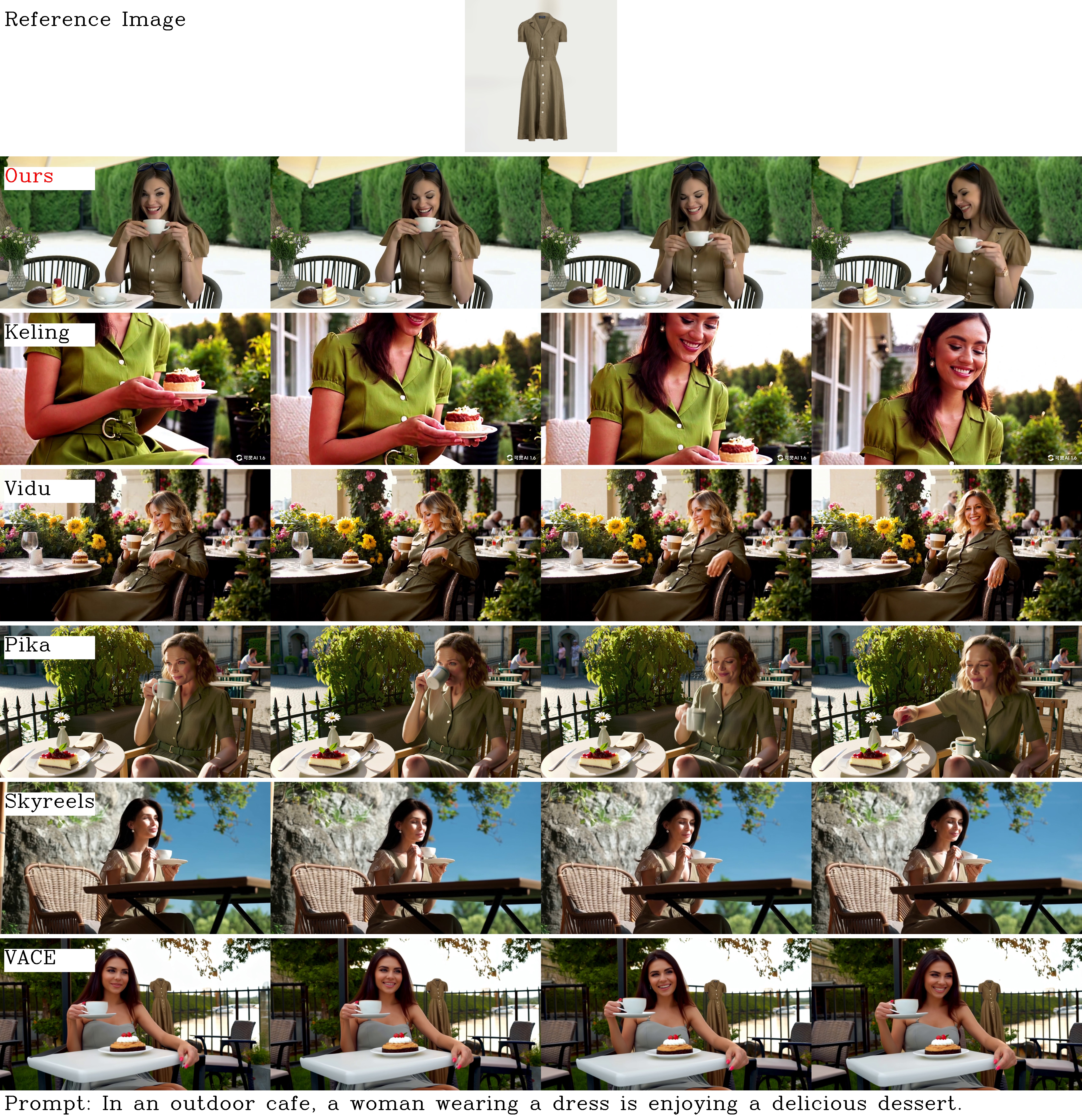}
    \caption{Comparison on object-centered video customization.}
    \label{fig:singleref_2}
\end{figure}

\textbf{Qualitative comparison.} We show the comparison between the state-of-the-art methods in Fig.~\ref{fig:singleref_1} and Fig.~\ref{fig:singleref_2}. It can be seen that Vidu~\citep{vidu}, Skyreels A2~\citep{fei2025skyreels}, and our method achieve relatively good results in prompt alignment and subject consistency, but our video quality is better than Vidu and Skyreels, thanks to the good video generation performance of our base model, i.e., Hunyuanvideo-13B~\citep{hunyuanvideo}. Among commercial products, although Keling~\citep{Keling} has a good video quality, the first frame of the video has a copy-paste problem (row 2 in \ref{fig:singleref_1}), and sometimes the subject moves too fast and blurs (row 2 in Fig.~\ref{fig:singleref_2}), leading a poor viewing experience. Pika~\citep{pika} performs poorly in consistency and is prone to subtitle problems. Hailuo~\citep{Hailuo} can only maintain ID consistency, but not full-body consistency. Among open-source methods, VACE~\citep{jiang2025vace} cannot maintain ID consistency (row 7 in Fig.~\ref{fig:singleref_1}). In comparison, HunyuanCustom can generate videos with high identity consistency, while keeping a good generation quality and diversity.

\begin{table}[t]
    \centering
    \caption{Model Performance Evaluation. We compare HunyuanCustom with state-of-the-art video customization methods across multiple metrics, including ID consistency (Face-Sim), subject similarity (DINO-Sim), text-video alignment (CLIP-B-T),  temporal consistency (Temp-Consis), and dynamic degree (DD). \textbf{Bold} and \underline{underline} represent optimal and sub-optimal results, respectively.}.
    \begin{tabular}{ccccccc}
        \toprule 
        Models & Face-Sim & CLIP-B-T & DINO-Sim & Temp-Consis & DD \\ \hline
        VACE-1.3B~\citep{jiang2025vace} & 0.204 & \underline{0.308} & 0.569 & \textbf{0.967} & 0.53  \\ 
        Skyreels~\citep{fei2025skyreels} & 0.402 & 0.295 & 0.579 & 0.942 & 0.72  \\ 
        Pika~\citep{pika} & 0.363 & 0.305 & 0.485 & 0.928 & \underline{0.89}  \\ 
        Vidu2.0~\citep{vidu} & 0.424 & 0.300 & 0.537 & \underline{0.961} & 0.43  \\ 
        Keling1.6~\citep{Keling} & 0.505 & 0.285 & \underline{0.580} & 0.914 & 0.78  \\ 
        Hailuo~\citep{Hailuo} & \underline{0.526} & \textbf{0.314} & 0.433 & 0.937 & \textbf{0.94} \\ 
        \textbf{HunyuanCustom (Ours)} & \textbf{0.627} & 0.306 & \textbf{0.593} & 0.958 & 0.71  \\
        \bottomrule 
    \end{tabular}
    \label{tab:singleref}
\end{table}

\textbf{Quantitative comparison.} We conduct a quantitative comparison between the state-of-the-art methods in Tab.~\ref{tab:singleref}. Our HunyuanCostom achieves the best ID consistency and subject consistency. It also achieves comparable results in prompt following and temporal consistency. Hailuo~\citep{Hailuo} has the best clip score because it can follow text instructions well with only ID consistency, sacrificing the consistency of non-human subjects (the worst DINO-Sim). In terms of Dynamic-degree, Vidu~\citep{vidu} and VACE~\citep{jiang2025vace} perform poorly, which may be due to the small size of the model.

 \begin{figure}[t]
    \centering
    \includegraphics[width=1.0\textwidth]{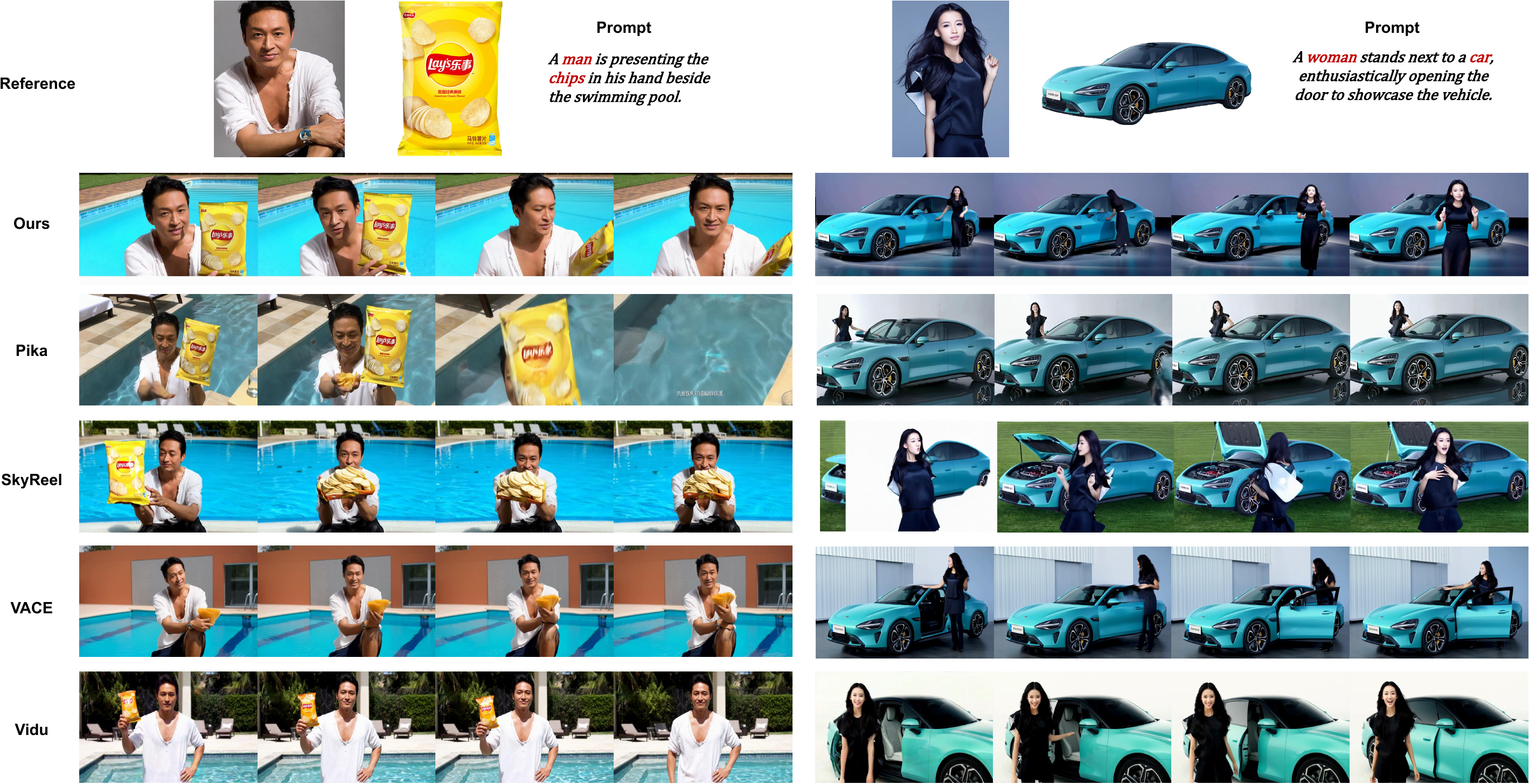}
    \caption{Comparison on multi-subject video customization.}
    \label{fig:multiref}
\end{figure}

 \begin{figure}[t]
    \centering
    \includegraphics[width=1.0\textwidth]{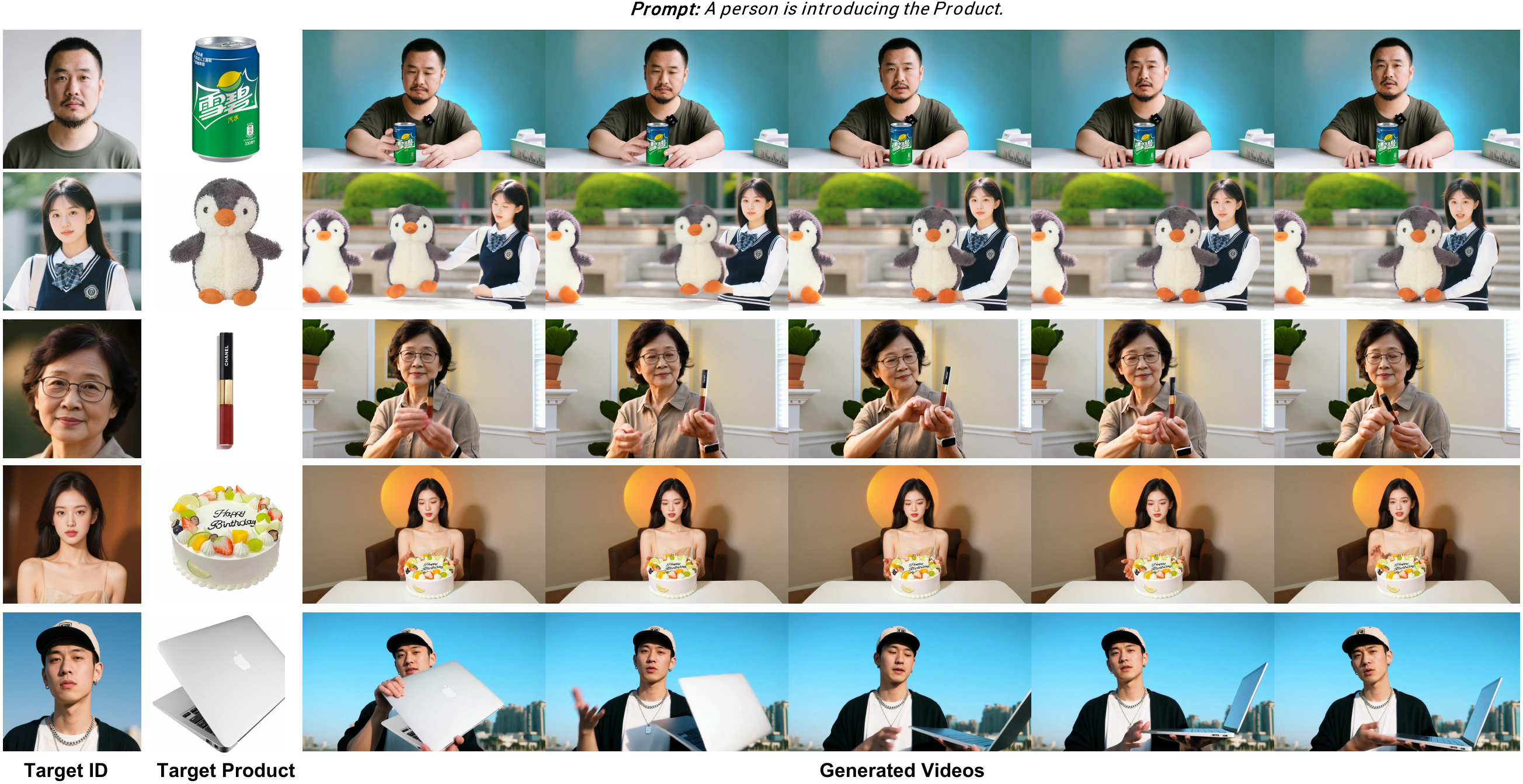}
    \caption{The results of our HunyuanCustom in virtual human advertisement, where HunyuanCustom can generate advertisement videos with good interaction between the human and products.}
    \label{fig:advertisement}
\end{figure}

\subsection{Experiments and Applications on multi-subject video customization}
\label{sec:multi-subject customization experiment}
 \textbf{Qualitative comparisons} In the context of multi-subject customization, we evaluate HunyuanCustom against leading multi-subject video customization methods, including commercial software such as Vidu 2.0~\citep{vidu}, Keling 1.6~\citep{Keling}, Pika~\citep{pika}, and Hailuo~\citep{Hailuo}, as well as open-source methods like VACE~\citep{jiang2025vace} and SkyReels A2~\citep{fei2025skyreels}. The comparative results are presented in Fig.~\ref{fig:multiref}. Pika~\citep{pika} can generate the specified subjects but exhibits instability in video frames, with instances of a man disappearing in one scenario and a woman failing to open a door as prompted. Vidu~\citep{vidu} and VACE~\citep{jiang2025vace} partially capture human identity but lose significant details of non-human objects, indicating a limitation in representing non-human subjects. SkyReels A2~\citep{fei2025skyreels} experiences severe frame instability, with noticeable changes in chips and numerous artifacts in the right scenario. In contrast, our HunyuanCustom effectively captures both human and non-human subject identities, generates videos that adhere to the given prompts, and maintains high visual quality and stability.

\textbf{Virtual Human Advertisement.} Leveraging our multi-subject customization capability, HunyuanCustom enables applications that previous methods cannot achieve. A significant application is in virtual human advertising, where HunyuanCustom takes a human image and a product image as inputs to generate a corresponding advertisement video. The results, shown in Fig.~\ref{fig:advertisement}, demonstrate that HunyuanCustom effectively maintains the identity of the human while preserving the details of the target product, including the text on it. Furthermore, the interaction between the human and the product appears natural, and the video adheres closely to the given prompt, highlighting the substantial potential of HunyuanCustom in generating advertisement videos.

 \begin{figure}[t]
    \centering
    \includegraphics[width=1.0\textwidth]{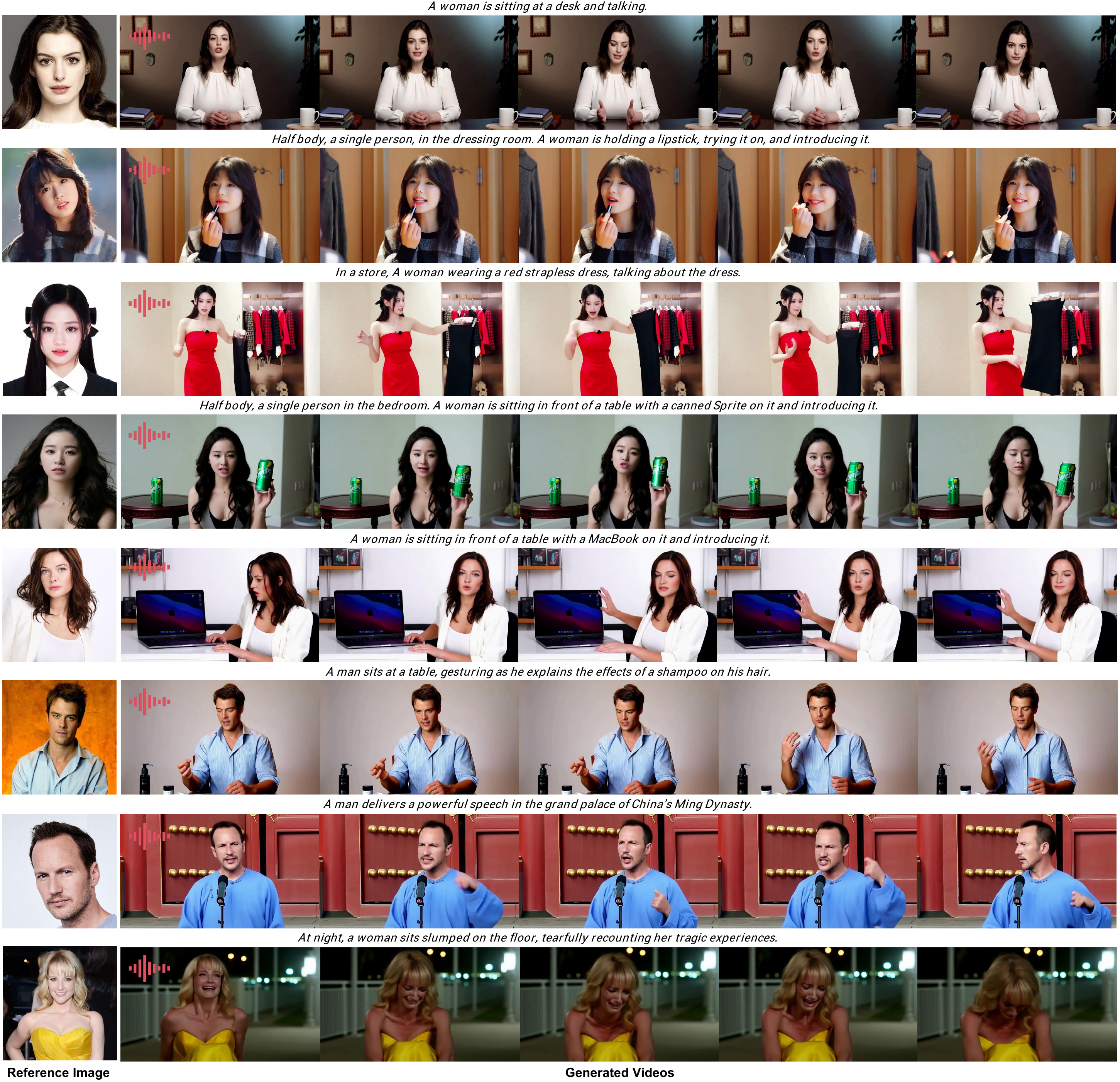}
    \caption{The results of our HunyuanCustom in Audio-driven customization, where we can generate videos in different scenes and postures specified by the text prompt, while keeping the identity well.}
    \label{fig:audio driven video}
\end{figure}

 \begin{figure}[t]
    \centering
    \includegraphics[width=1.0\textwidth]{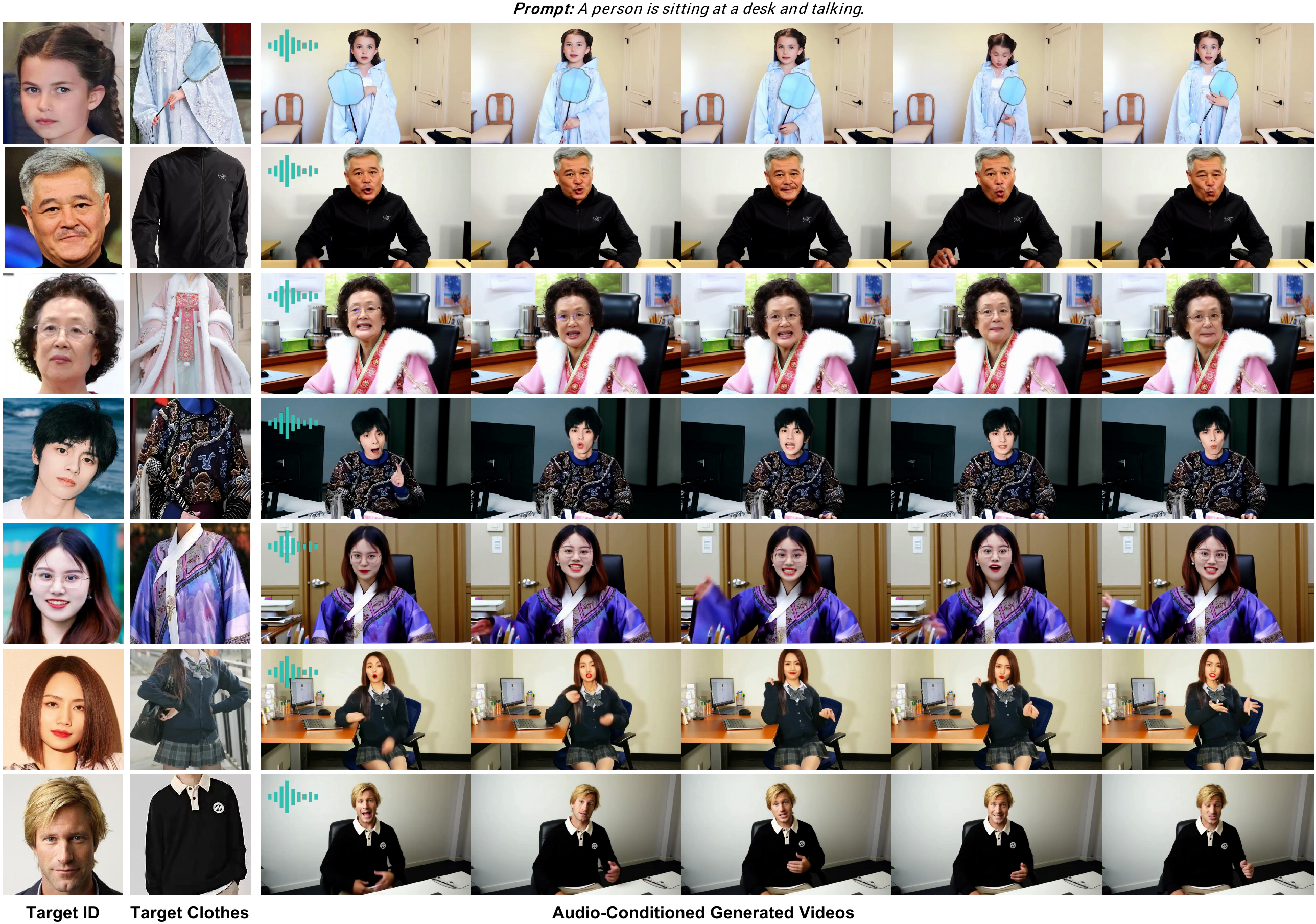}
    \caption{The results of our HunyuanCustom in audio-driven multi-subject customization, where we can generate humans in different clothes, while speaking the given audio vividly.}
    \label{fig:multi-subject audio driven video}
\end{figure}

\subsection{Experiments on audio-driven video customization}

\textbf{Audio-driven single-subject customization.} Previous audio-driven human animation methods input a human image and an audio, where the human posture, attire, and environment remain consistent with the given image and cannot generate videos in other gestures and environments, which may restrict their application.  In comparison, our HunyuanCustom enables audio-driven human customization, where the character speaks the corresponding audio in a text-described scene and posture, allowing for more flexible and controllable audio-driven human animation. The generated results are illustrated in Fig.~\ref{fig:audio driven video}. HunyuanCustom produces videos that closely align with the given prompts while preserving character identities. It demonstrates effective interaction with other subjects (rows 3 \& 4) or humans (rows 5 \& 6), which can significantly enhance its application in live streaming and advertising. Additionally, it can generate videos featuring diverse scenes and postures, such as those set in the Ming Dynasty (row 7), where characters are automatically dressed in period-appropriate attire without explicit prompts, and row 8 showcases a woman with vivid and realistic expressions distinct from the input image. This demonstrates HunyuanCustom's robust world modeling and generalization capabilities. In summary, our audio-driven HunyuanCustom can generate videos across various scenes and postures specified by text prompts with high diversity, while keeping the identity well.

\textbf{Audio-driven virtual try-on.} Utilizing its multi-subject customization capability, HunyuanCustom also supports audio-driven multi-subject video customization, offering a wide range of applications. In Sec~\ref{sec:multi-subject customization experiment}, we demonstrated HunyuanCustom's capabilities in virtual human advertising. Here, we further explore its generation ability in virtual try-on, driven by both text prompts and audios. The results, shown in Fig.~\ref{fig:multi-subject audio driven video}, illustrate the integration of virtual try-on with audio-driven video generation. The generated videos effectively preserve the target identities while naturally the specified attire and synchronizing vividly with the given audio. This highlights HunyuanCustom's robust capability in multi-modal video customization.

 \begin{figure}[t]
    \centering
    \includegraphics[width=1.0\textwidth]{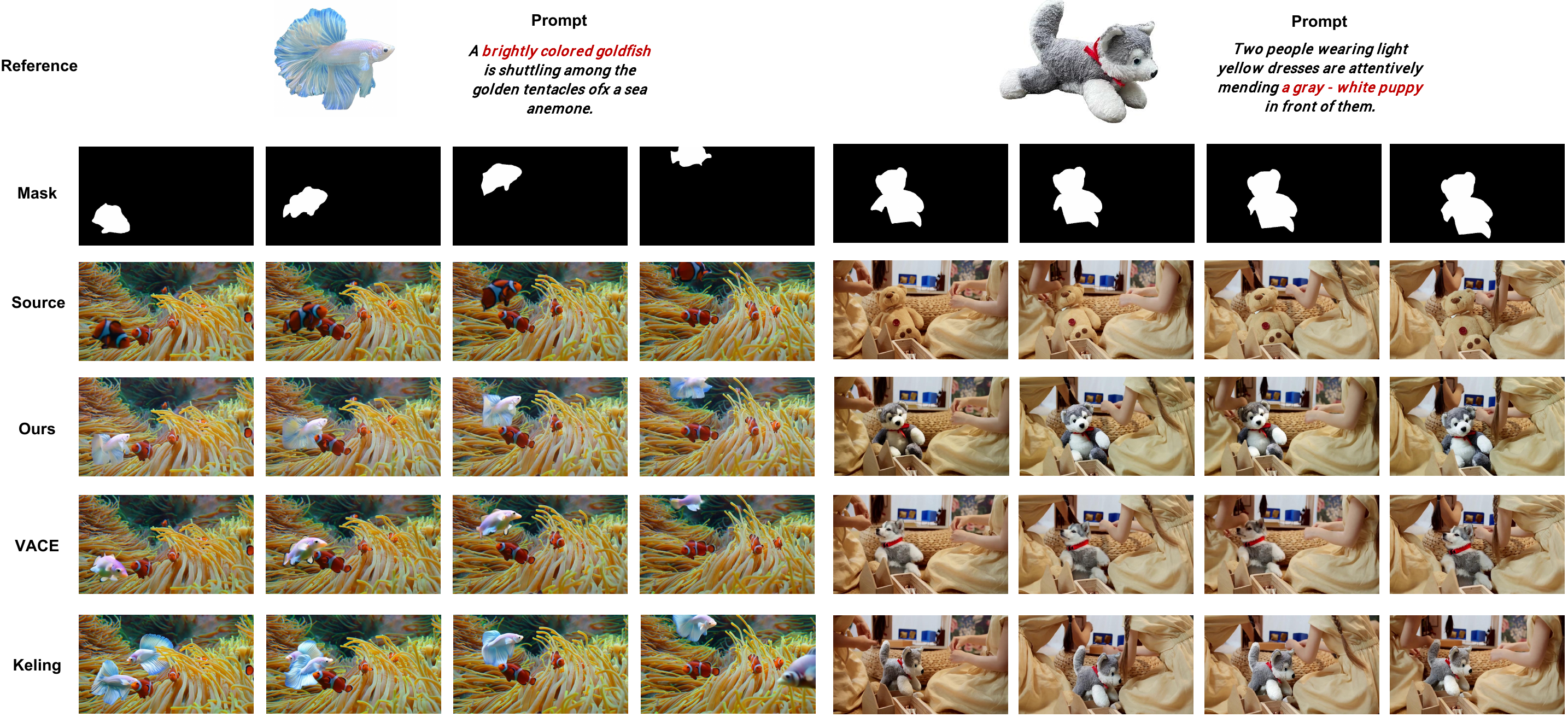}
    \caption{The results of our HunyuanCustom in Video-driven video customization, where we can edit anything in the source video with a given mask video, while generating video vividly.}
    \label{fig:edit}
\end{figure}

\subsection{Experiments on video-driven video customization}
\textbf{Video subject replacement.}
Leveraging its strong subject consistency, HunyuanCustom also supports video-driven video editing, enabling a broad range of application scenarios. We compare HunyuanCustom with VACE~\cite{jiang2025vace} and Keling~\cite{Keling} on the task of video subject replacement, where a source video, object masks indicating regions to be replaced, and a target subject image are provided as inputs. The results are presented in Fig.~\ref{fig:edit}. VACE suffers from boundary artifacts due to strict adherence to the input masks, resulting in unnatural subject shapes and disrupted motion continuity. Keling, in contrast, exhibits a copy-paste effect, where subjects are directly overlaid onto the video, leading to poor integration with the background. In comparison, HunyuanCustom effectively avoids boundary artifacts, achieves seamless integration with the video background, and maintains strong identity preservation—demonstrating its superior performance in video editing tasks.

 \begin{figure}[t]
    \centering
    \includegraphics[width=1.0\textwidth]{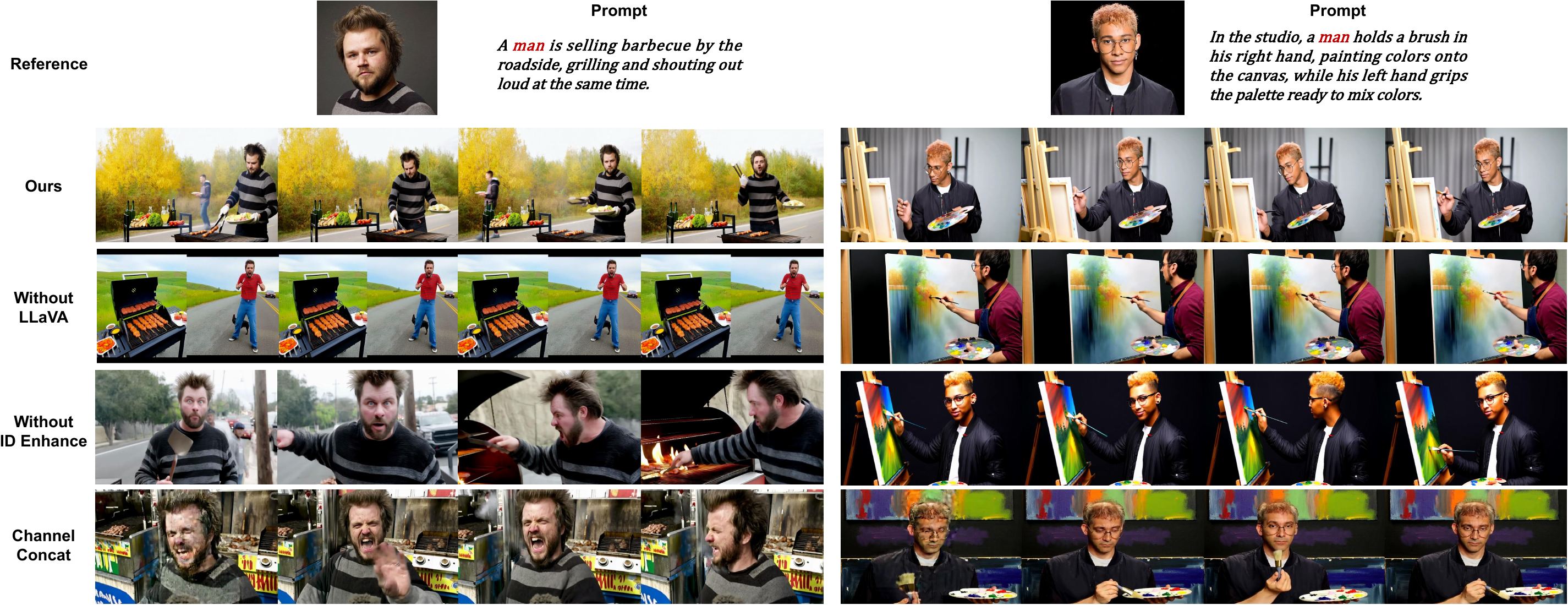}
    \caption{Ablation study on the proposed modules in HunyuanCustom.}
    \label{fig:ablation study}
\end{figure}

\subsection{Ablation Study}

We conduct ablation studies on subject customization, where we compare with three ablated models: (1) the model without LLaVA; (2) the model without identity enhancement; (3) the model with identity enhancement by channel-level concatenation. The results, presented in Fig.~\ref{fig:ablation study}, reveal that the model without LLaVA exhibits poor identity preservation, indicating that LLaVA not only conveys prompt information but also extracts key identity features. The model without LLaVA fails to capture any significant details from the target image. Additionally, the model with LLaVA but lacking identity enhancement captures global identity information but misses detailed identity features, demonstrating the effectiveness of the identity enhancement module in refining identity details. Finally, the model using channel concatenation instead of temporal concatenation shows poor generation quality. Although it captures identity well, it suffers from a severe blurring effect in the initial frames, similar to results from Vidu~\citep{vidu}. This suggests that temporal concatenation aids in effectively capturing target information through strong temporal modeling priors and minimizes the impact on generation quality. In summary, our model successfully captures both global and local identity details while ensuring high generation quality, underscoring the effectiveness of our design.

%% file: secs/5_conclusion.tex
\section{Conclusion}
\label{sec:Conclusion}
In this work, we propose HunyuanCustom, a novel multi-modal customized video generation model that addresses the critical challenge of subject-consistent video generation and enables multi-modal identity-centric video customization. By combining image, audio, and video modalities with a text-driven conditioning mechanism, HunyuanCustom offers a robust framework for generating high-quality videos with precise identity consistency. Our model's integration of a text-image fusion module, an image ID enhancement module, and an efficient audio and video feature injection process ensures that the generated videos adhere to the user’s specific requirements, achieving both high fidelity and flexibility.

Through extensive experiments, we have demonstrated that HunyuanCustom outperforms existing methods in various tasks, including single- and multi-subject generation, audio-driven and video-driven video customization. The results show superior performance in terms of ID consistency, authenticity, and video-text alignment, positioning HunyuanCustom as a leading solution for controllable video customization. This work paves the way for future research in controllable video generation, further expanding the potential applications of Artificial Intelligence Generated Content (AIGC) in creative industries and beyond.

\section{Contributors and Ackonwledgements}

\begin{itemize}
    \item \textbf{Project leaders:} Qinglin Lu, Qin Lin, Yuan Zhou;
    \item \textbf{Core Contributors:} Zhentao Yu, Zhengguang Zhou, Teng Hu, Sen Liang;

    \item \textbf{Acknowledgements:}
    We would like to thank Yi Chen, Zixiang Zhou, Hongmei Wang, Yuanbo Peng,  Zunnan Xu, Linqing Wang, Yifu Sun, Sihuan Lin for their valuable inputs and suggestions.

\end{itemize}